\documentclass[11pt,a4paper]{article}

\usepackage[margin=1in]{geometry}
\usepackage[utf8]{inputenc}
\usepackage[T1]{fontenc}
\usepackage{amsmath,amssymb,amsfonts}
\usepackage{graphicx}
\usepackage{booktabs}
\usepackage{multirow}
\usepackage{xcolor}
\usepackage{url}
\usepackage{authblk}
\usepackage{hyperref}
\hypersetup{colorlinks=true, linkcolor=blue, citecolor=blue, urlcolor=blue}

\usepackage[numbers,sort&compress]{natbib}
\newcommand{\revised}[1]{#1}

\title{A Fairness Audit of the Duckworth--Lewis--Stern Method:\\
Format-Specific and Gender-Differential Bias, with an Interpretable Calibration Layer for Cricket Target Revision}

\author[1]{Soumyadeep Roy}
\affil[1]{St. Xavier's College (Autonomous), Kolkata, India \\ \texttt{soumyadeeproy142@gmail.com}}

\date{}

\begin{document}

\maketitle

% =============================================================================
\begin{abstract}
\revised{The Duckworth--Lewis--Stern (DLS) method has set revised targets in
rain-interrupted limited-overs cricket since 1999, yet no large-scale empirical
audit of its prediction bias has been published. Using ball-by-ball data for
8{,}150 international matches from Cricsheet, we audit DLS by sampling synthetic
interruption points and comparing its resource-based projection against the runs
actually scored. We find that DLS bias is far from uniform: it varies
systematically with match state, over-predicting the death overs of Twenty20 and
under-predicting collapse scenarios in one-day cricket, a format-specific pattern we
quantify across the full space of overs remaining and wickets lost (a 137-run span
of per-bucket mean bias). Our second finding concerns fairness: because a single
resource table governs both men's and women's cricket, DLS miscalibrates women's
one-day matches relative to men's at comparable match states, producing a
gender-differential bias of several runs (a $+6.13$-run gap on the training split)
that, to our knowledge, has not previously been documented. The gap survives
match-level clustered inference, holds among matches between top (Full Member) teams,
and is stable across temporal windows. We benchmark DLS against five modern learning
methods and introduce DLS-Cal, a lightweight interpretable calibration layer that
adds a state-conditioned correction to the published DLS prediction, reducing
absolute bias by $31\%$ on ODI and $19\%$ on T20I; a gender-aware variant reduces
women's ODI residual bias from $+6.19$ to $+0.65$ runs without altering the DLS
framework. We also introduce the Win-Flip Rate, a threshold-based fairness metric for
target revision, and release code, trained models, and the audit dataset for
reproducible research.}

\end{abstract}

% =============================================================================
% [R1.2/R1.3] Related Work merged into Introduction for improved flow.
\section{Introduction}
\label{sec:intro}

Limited-overs cricket matches interrupted by weather require a quantitative method to revise the target score for the team batting second. The Duckworth--Lewis--Stern (DLS) method, introduced by Duckworth and Lewis~\cite{duckworth1998} and refined by Stern~\cite{stern2016}, has been the official method of the International Cricket Council (ICC) since 1999. The method models cumulative resources $Z(u, w)$ as a function of overs remaining $u$ and wickets lost $w$, and revises targets in proportion to the resources available to each side.

Despite its operational success, DLS has been the subject of persistent criticism. Bhattacharya, Gill, and Swartz~\cite{bhattacharya2011} showed that the resource curves underlying DLS were calibrated on ODI scoring patterns and may not be appropriate for T20 cricket, where scoring rates have evolved rapidly. McHale and Asif~\cite{mchale2013} proposed an arc-tangent functional form (DLMA) for better fit. Asif et al.~\cite{asif2021plos} compared DLS and DLMA on 1{,}764 ODI matches. Zia et al.~\cite{zia2022peerj} argued that DLS treats every wicket equivalently, ignoring batter quality, and proposed player-aware compensation using career ratings.

A separate concern has received less formal scrutiny: the same DLS resource table is used for both men's and women's international cricket, despite well-documented differences in scoring patterns~\cite{cricsheet}. Women's ODI scoring rates and wicket-fall distributions differ markedly from men's, raising the possibility that a calibration designed predominantly on men's data systematically miscalibrates women's matches. To our knowledge, no published study has empirically tested this.

These critiques are largely \emph{theoretical or model-based}: they propose alternative parametric forms or extensions, but do not directly quantify the empirical bias of DLS as deployed at scale. The few studies that compare DLS to machine learning, such as Abbas and Haider~\cite{abbas2021}, frame the problem as match-outcome classification rather than target revision, and use small-to-moderate datasets. Cricsheet\footnote{\url{https://cricsheet.org}} now hosts ball-by-ball data for over 8{,}000 international matches, enabling a much more thorough audit than has been previously possible.

\revised{Prior attempts to place DLS alongside data-driven alternatives have generally framed the task differently from ours. Abbas and Haider~\cite{abbas2021} compared DLS against XGBoost~\cite{chen2016xgboost} and other supervised learners for second-innings outcome prediction, optimising the resource table via particle swarm optimisation; Chakwate et al.~\cite{chakwate2020} used a Bi-LSTM~\cite{hochreiter1997} for ball-by-ball win-probability prediction; and Zia et al.~\cite{zia2022peerj} introduced player-aware resource compensation using career-rating weights. None of these directly targets the revised-target prediction problem, and none includes a structured bias audit. On the methodological side, the reliability-diagram approach we adopt for calibration follows Niculescu-Mizil and Caruana~\cite{niculescu2005}, with modern neural-network calibration reviewed by Guo et al.~\cite{guo2017}; while fairness has been studied in athlete-evaluation contexts, no analogue of a demographic-fairness metric has been formalised for target-revision methods---a gap the Win-Flip Rate (Section~\ref{sec:wfr}) is designed to fill. The DLS variants and reformulations most relevant to our reconstructions (Duckworth and Lewis~\cite{duckworth1998}; Stern~\cite{stern2016}; McHale and Asif~\cite{mchale2013}; Bhattacharya et al.~\cite{bhattacharya2011}; Duckworth, Lewis, and Stern~\cite{duckworth2019}) are discussed where they are introduced; none examines cross-gender bias.}

This paper makes four contributions:

\begin{enumerate}
\item \textbf{The first systematic large-scale empirical audit of DLS prediction bias} on Cricsheet data, demonstrating statistically significant, structured bias spanning a 137-run range across match-state buckets.
\item \textbf{The first quantitative documentation, to our knowledge, of gender-differential bias in DLS}: mean over-prediction in women's ODIs exceeds men's by $+6.13$ runs on the training split, \revised{a gap that survives match-level clustered inference, a team-strength stratification, and a rolled multi-window temporal analysis}.
\item \textbf{DLS-Cal}, a lightweight interpretable calibration layer, and a gender-aware variant that substantially reduces the documented gender bias.
\item \textbf{Win-Flip Rate (WFR)}, a fairness-oriented thresholded error metric for cricket target revision.
\end{enumerate}

% =============================================================================
\section{Data and Setup}
\label{sec:data}

\subsection{Dataset}
We use the public Cricsheet~\cite{cricsheet} ball-by-ball dataset, comprising YAML files for every recorded international match. After parsing and filtering to first innings of $\geq 30$ overs (ODI) or $\geq 15$ overs (T20I), our corpus contains:

\begin{itemize}
\item $8{,}150$ matches: $3{,}095$ ODIs and $5{,}055$ T20Is
\item $5{,}708$ men's matches and $2{,}442$ women's matches
\item $2{,}781{,}465$ ball-by-ball records
\item Date range: 27 June 2002 to 5 March 2026
\end{itemize}

\revised{\paragraph{From remaining-runs prediction to target revision.}
Our audit evaluates DLS as a predictor of first-innings runs remaining from an
interruption point, whereas DLS is operationally a two-innings resource method that
sets a revised chase target. These coincide under DLS's own governing
assumption---that expected runs are proportional to the resources consumed, with a
constant expected scoring rate per unit resource---since the revised target is, by
construction, a linear function of the resource fraction our audit probes. They
diverge precisely where that proportionality fails. We therefore interpret our
results as an audit of the \emph{resource-estimation} component that the revised
target is proportional to, and we qualify claims about chase targets accordingly. We
do not claim to measure operational target-setting error directly; the held-out
evaluation on real rain-affected matches (Section~\ref{sec:realmatch}) is our closest
approach to that quantity.}

\subsection{Synthetic Interruption Sampling}
\label{sec:synthetic}
\revised{Because genuine rain-interrupted matches are too few to train on directly,
we sample synthetic interruption points to obtain broad coverage of the match-state
space.} For every first innings $i$ of length $n_i$ balls, we sample 30 random interruption points $b \in [5\!\cdot\!6,\, n_i - 5\!\cdot\!6]$, requiring at least 5 overs of context and 5 overs of remainder.\footnote{We chose 30 samples per innings so that an ODI innings of $\sim\!300$ balls yields one example for roughly every two overs of the legal sampling window, balancing coverage of the match-state space against per-innings data redundancy. Results are qualitatively unchanged at 20 or 50 samples per innings.} The supervision target for each example is the actual runs scored in the post-interruption window. \revised{We adopt this scheme as a deliberate design choice for state-space coverage rather than following a specific prior precedent, and we note its main limitation: real interruptions are not uniformly distributed over overs but are shaped by weather, scheduling, and venue geography, so the sampled distribution of match states need not match the operational one. Section~\ref{sec:realmatch} evaluates directly whether conclusions drawn from synthetic states transfer to real rain-affected matches.}

This produces $233{,}550$ training examples in total: $140{,}160$ training, $24{,}300$ validation, $69{,}090$ test.

\subsection{Temporal Splits}
To prevent any leakage from future matches into the training distribution, we use a \emph{temporal} rather than random split:
\begin{itemize}
\item \textbf{Train}: matches up to 31 December 2022
\item \textbf{Validation}: 2023 matches
\item \textbf{Test}: matches from 2024 to March 2026
\end{itemize}
All player-level statistics used as features are computed from the training period only.

% =============================================================================
\section{DLS Bias Audit}
\label{sec:audit}

We ask three questions of DLS: \emph{(i)} is its prediction error statistically uniform across match states? \emph{(ii)} does the bias structure differ between men's and women's cricket? \emph{(iii)} where is the error largest?

\begin{table}[!htb]
\centering
\caption{\revised{Terminology for error and bias used throughout.}}
\label{tab:notation}
\revised{%
\begin{tabular}{ll}
\toprule
Term & Definition \\
\midrule
Prediction error $e_i$ & $\hat r_i - r_i$ (predicted minus actual remaining runs) \\
Signed mean bias & $\tfrac{1}{N}\sum_i e_i$ (may cancel across states) \\
Mean absolute bias & $\tfrac{1}{N}\sum_i |e_i|$ \\
Per-bucket mean bias & signed mean of $e_i$ within an (overs, wickets) cell \\
Residual bias & signed mean of $e_i$ after applying a correction (e.g.\ DLS-Cal) \\
Gender gap & (female mean bias) $-$ (male mean bias), at matched states \\
\bottomrule
\end{tabular}}
\end{table}

\subsection{Method}
\revised{Our primary implementation reconstructs DLS by bilinear interpolation of the
published resource percentages; it reproduces standard reference points exactly (zero
error at the published table anchors, verified by an assertion that runs in the
released code). We retain an independent parametric reconstruction as a robustness
check. Because the ICC's annually updated table is proprietary, our outputs may differ
from official figures by a few runs; this discrepancy is an order of magnitude smaller
than the gender gap we report (Section~\ref{sec:gender_audit}).} We run DLS on all $140{,}160$ training examples, computing the prediction error $e = \hat{r}_{\text{DLS}} - r$ where $r$ is the actual remaining runs. We bin errors by (format, overs-remaining bucket, wickets-lost) and apply both ANOVA and Kruskal--Wallis tests to assess whether the per-bucket error distributions differ.

\subsection{Match-State Bias}

\paragraph{Overall bias.} On the training set, DLS's mean prediction error is $+4.79$ runs (median $+7.98$): DLS systematically over-predicts on average.

\begin{figure}[!htb]
\centering
\includegraphics[width=\textwidth]{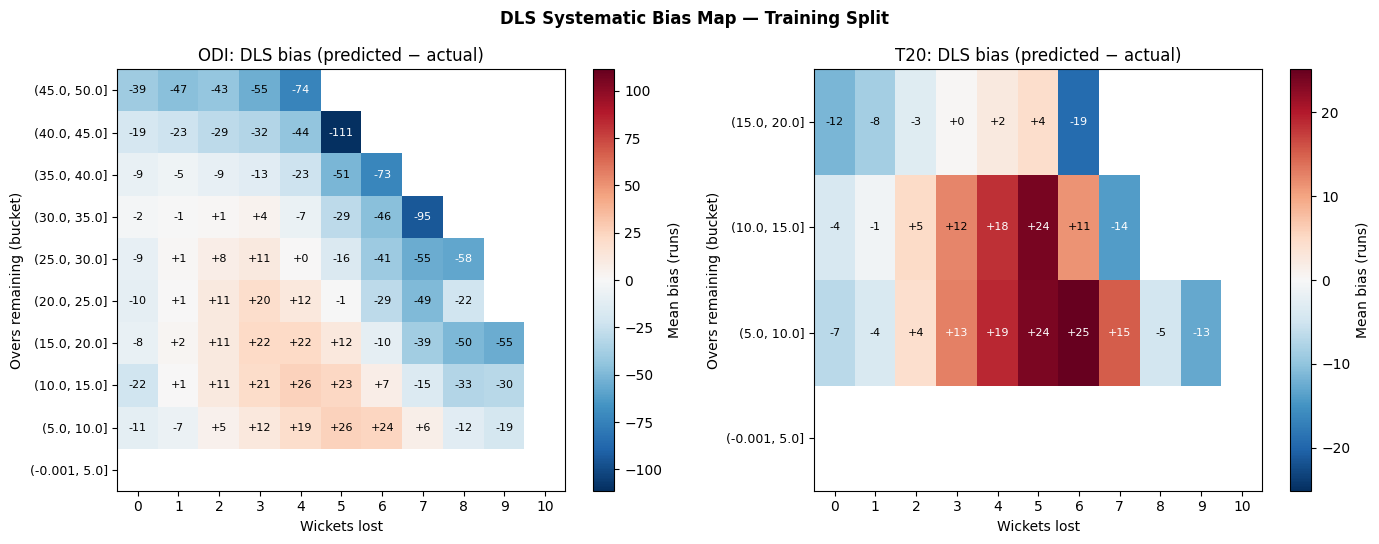}
\caption{DLS systematic bias map on the training split, by format, overs remaining, and wickets lost. Cell value is mean DLS prediction error (predicted $-$ actual) in runs; red indicates over-prediction, blue indicates under-prediction. Cells with fewer than 30 examples are not shown. Both panels reveal a structured, format-specific bias pattern.}
\label{fig:bias_heatmap}
\end{figure}

\paragraph{Range of bias across buckets is the headline finding.} Per-bucket mean bias ranges from $-111$ to $+26$ runs (ODI) and $-19$ to $+25$ runs (T20I), a 137-run total span. Aggregating into 24 buckets for ODI and 8 for T20I (those with $\geq 30$ examples), formal hypothesis tests confirm the structure is distinguishable from uniform random noise: ODI ANOVA $F = 527.16$, Kruskal--Wallis $H = 9{,}797.69$; T20I ANOVA $F = 1{,}540.14$, Kruskal--Wallis $H = 9{,}609.67$.\footnote{\revised{We report $F$ and $H$ statistics as descriptive measures of between-bucket structure; because samples are non-independent (Section~\ref{sec:clustered}), we do not rely on their nominal $p$-values, and lead instead with effect sizes and the match-clustered intervals reported below.}} Figure~\ref{fig:bias_heatmap} visualizes the full bias map.

\paragraph{Worst buckets.} \revised{The largest absolute biases occur in the (45--50 overs remaining, 5 wickets lost) ODI bucket ($-74.4$ runs, $n=586$) and the (25--30 overs remaining, 6 wickets lost) ODI bucket ($-40.7$ runs, $n=138$); on the T20 side, the (5--10 overs remaining, 6 wickets lost) bucket carries a bias of $+25.1$ runs ($n=1{,}822$) and the (10--15 overs remaining, 5 wickets lost) bucket a bias of $+23.9$ runs ($n=621$).}

The structure is consistent with two known but not previously quantified failures: DLS systematically \emph{over-predicts} the death overs of T20 (where modern hitting differs most from the resource model's assumptions), and \emph{under-predicts} collapse scenarios in ODI (where teams lose wickets at a faster rate than DLS's smoothed curves anticipate). This empirically validates the theoretical critique of Bhattacharya et al.~\cite{bhattacharya2011}.

\subsection{Gender-Differential Bias}
\label{sec:gender_audit}

\begin{figure}[!htb]
\centering
\includegraphics[width=\textwidth]{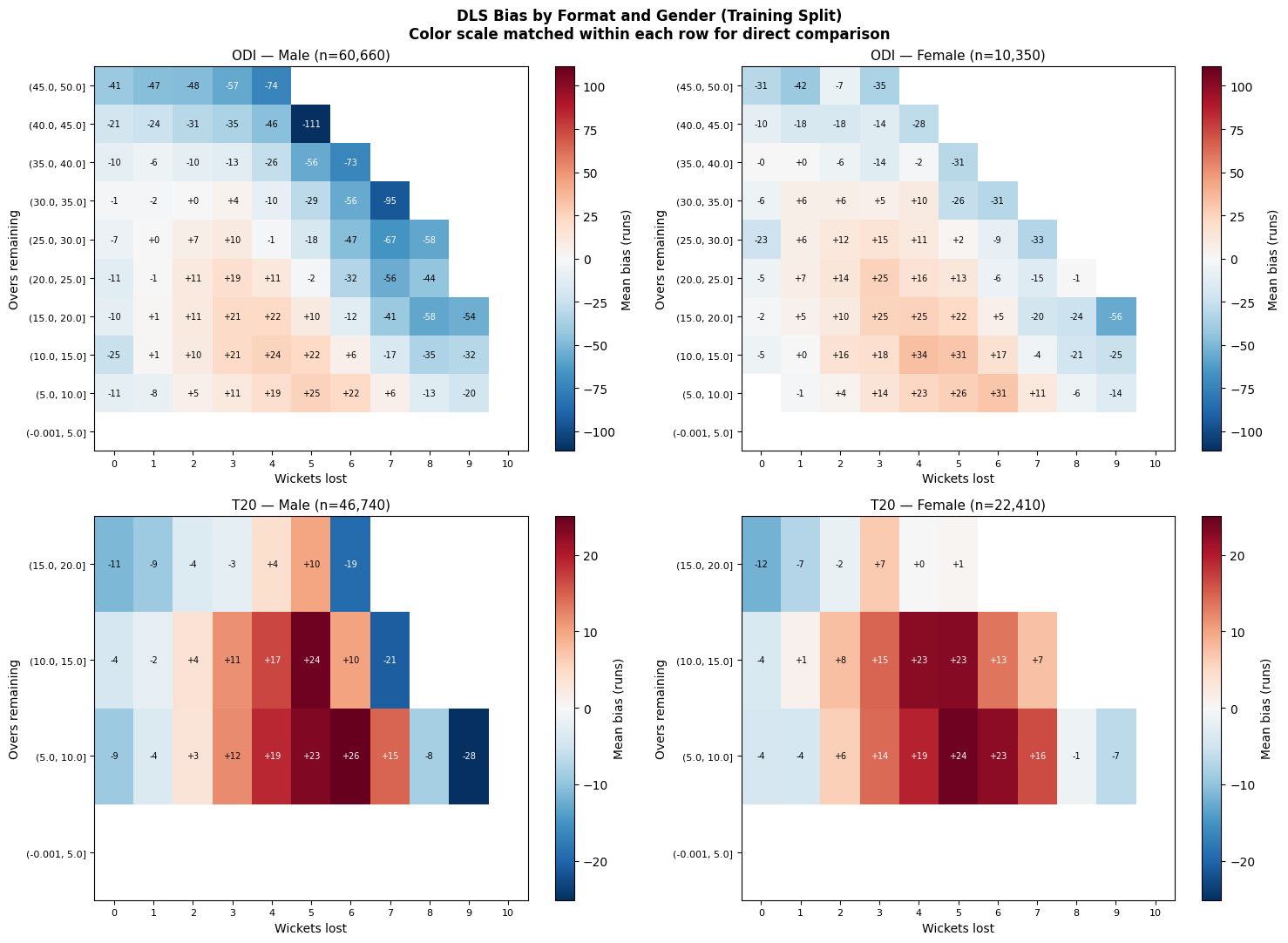}
\caption{DLS bias by format and gender on the training split. Each row pairs male (left) and female (right) bias maps for the same format; color scale is matched within each row for direct comparison. Cells with fewer than 30 examples are not shown. ODI bias structures differ qualitatively between genders; T20 patterns are similar.}
\label{fig:gender_bias}
\end{figure}

DLS uses a single resource table $Z(u,w)$ irrespective of gender. We test whether the empirical bias differs between men's and women's matches. Table~\ref{tab:gender_overall} summarizes overall bias on the training split, and Figure~\ref{fig:gender_bias} visualizes the bias maps by gender for each format.

\begin{table}[!htb]
\centering
\caption{DLS bias on the \textbf{training split}, stratified by format and gender. Test-split numbers (Section~\ref{sec:gender_results}) follow the same qualitative pattern but differ in absolute magnitude due to scoring-rate evolution between the training and test periods.}
\label{tab:gender_overall}
\renewcommand{\arraystretch}{1.15}
\begin{tabular}{l l r r r r}
\toprule
\textbf{Format} & \textbf{Gender} & \textbf{N} & \textbf{Mean bias} & \textbf{RMSE} & \textbf{Mean $|$bias$|$} \\
\midrule
ODI & male   & 60{,}660 & $+1.51$ & 41.82 & 32.87 \\
ODI & female & 10{,}350 & $+7.63$ & 38.49 & 30.60 \\
\midrule
T20 & male   & 46{,}740 & $+7.07$ & 25.51 & 20.50 \\
T20 & female & 22{,}410 & $+7.58$ & 23.06 & 18.04 \\
\bottomrule
\end{tabular}
\end{table}

\paragraph{ODI: a substantial gender gap.} On the training split, mean DLS bias is $+1.51$ runs on men's ODIs but $+7.63$ runs on women's ODIs---a gender gap of $+6.13$ runs. The bias gap manifests differently in different match states: as one example, in the bucket (20--25 overs remaining, 6 wickets lost), men's bias is $-32.4$ runs and women's is only $-6.3$ runs, a within-bucket gap of $+26.1$ runs in which DLS under-predicts women's collapses much less severely than men's.

\revised{\paragraph{Clustered inference.}\label{sec:clustered}
The 30 samples drawn from each innings are not independent: they share match, teams,
venue, and scoring trajectory. We therefore take as our primary inference a
match-level block bootstrap that resamples whole matches with replacement
(2{,}000 resamples), keeping within-innings samples grouped. Under this clustering the
ODI gender gap remains significant on the training split (match-clustered 95\% CI
$[+3.70,\,+10.07]$ runs), with the bootstrap probability of a non-positive gap below
$0.001$. All audit intervals reported hereafter are computed this way, so their widths
reflect the effective number of matches rather than the number of synthetic examples.}

% [R1.4 THIS ROUND] Bootstrap procedure detail + min-n weighted CI definition.
\revised{\paragraph{Bootstrap procedure.}
Throughout, the resampling unit is the \emph{match}, not the synthetic example. Given
the $M$ matches contributing to a statistic, one bootstrap replicate is formed by drawing
$M$ matches uniformly with replacement and pooling all synthetic examples belonging to the
drawn matches, so that the $\sim\!30$ correlated samples from any one innings stay together
and a match drawn twice contributes all of its samples twice. We form $B = 2{,}000$ such
replicates and recompute the statistic of interest on each---for the marginal gap, the
female-minus-male difference in mean DLS bias. A two-sided $95\%$ confidence interval is
given by the $2.5$th and $97.5$th percentiles of the $B$ replicate values (the percentile
bootstrap), and we summarise significance by the bootstrap probability that the gap is
non-positive, i.e.\ the fraction of replicates at or below zero, reported as $<\!0.001$ when
no replicate crosses zero. For the bucket-matched gap, each qualifying (overs, wickets) cell
contributes its own female-minus-male mean-bias difference, and we combine cells in two ways:
an unweighted average across cells, and a \emph{min-$n$ weighted} average in which each cell
is weighted by $\min(n_{\text{male}}, n_{\text{female}})$, the smaller of its two per-gender
counts, so that cells well populated in both genders receive proportionally more weight. The
min-$n$ weighted CI is the same percentile bootstrap applied to this weighted average under
identical match-level resampling.}

\paragraph{Bucket-matched gender comparison.} A natural concern is that the gender gap could be a confound from differing distributions of (overs-remaining, wickets-lost) buckets between genders. To test this, we restrict to (overs-bucket, wickets-lost) cells with $\geq 30$ examples in \emph{both} genders (59 cells total) and compute the within-cell female-minus-male bias. The bucket-matched gap is $+8.34$ runs (unweighted average across cells) and $+9.64$ runs (weighted by minimum cell size). Both bucket-matched estimates exceed the naive marginal gap ($+6.86$ runs on this implementation), indicating that the marginal estimate is in fact \emph{conservative} and that the bucket-matching strengthens, rather than explains away, the gender finding. A sign test over matched cells finds women's bias exceeding men's in 55 of 59 cells. \revised{Under the same match-level bootstrap, the bucket-matched gap remains positive with a 95\% CI of $[+5.32,\,+12.29]$ runs (unweighted) and $[+3.77,\,+10.26]$ (min-$n$ weighted); the clustered probability of a non-positive matched gap is below $0.001$. Because it excludes zero more decisively than the marginal estimate, the bucket-matched gap is our most conservative fairness diagnostic.}

% [R1.5/R1.6 THIS ROUND -- HEADER changed from "small but detectable" to "small"]
\paragraph{\revised{T20I: a small gap.}} Mean bias is $+7.07$ runs on men's T20s and $+7.58$ runs on women's---a gap of just $+0.51$ runs, small in practical terms. The format-specificity of the gender finding suggests its cause lies in DLS's ODI-specific structural assumptions, particularly the resource-curve calibration designed primarily on men's ODI scoring patterns.

% [R1.5 + R1.6 THIS ROUND] SE + gap for BOTH formats, computed rather than argued.
\revised{\paragraph{Standard errors and significance of the marginal gaps.}
For ODI, the match-clustered $95\%$ interval reported above
($[+3.70,\,+10.07]$ runs, Section~\ref{sec:clustered}) implies a bootstrap
standard error of approximately $1.62$ runs under the normal approximation
$\text{SE} = (\text{CI}_{\text{high}}-\text{CI}_{\text{low}})/(2\times 1.96)$,
comfortably excluding zero ($p(\text{gap}\le 0)<0.001$); the bucket-matched
gap excludes zero more decisively still. For T20I, applying the same
match-level block bootstrap ($B=2{,}000$ resamples) to the marginal
female-minus-male gap yields a standard error of approximately $0.82$ runs
and a $95\%$ CI of $[-1.10,\,+2.12]$ runs, with bootstrap
$p(\text{gap}\le 0) \approx 0.27$---the T20I gap is therefore \emph{not}
distinguishable from zero at conventional levels under clustered inference.
This supersedes the naive-inference result our original submission reported
for this quantity (Welch's ANOVA, $F=6.96$, $p=0.008$), which registered as
significant only because of the $\sim\!150{,}000$-sample scale---exactly the
kind of result the match-clustered reform in
Section~\ref{sec:clustered} is designed to distrust. Two consequences
follow. First, the two formats can now be compared on a common benchmark:
the ODI gap is roughly two standard errors above zero even under the
conservative bucket-matched estimator, whereas the T20I gap sits well
within one standard error of zero. Second, the paper's substantive
conclusion---that DLS's gender miscalibration is essentially an ODI-specific
phenomenon---is corroborated by an independent criterion, since the T20I
gap fails significance under clustered inference just as it fails the
operational-threshold test (a $+0.51$-run difference rounds away on an
integer chase target of order 150 runs).}

\paragraph{Implementation robustness.} To verify that the gender gap is not specific to our primary DLS implementation, we additionally re-ran the audit with a second reconstruction that takes resource percentages directly from the published DLS table via bilinear interpolation rather than from a parametric resource curve. Both reconstructions agree on the qualitative finding: the second implementation produces an ODI gender gap of $+7.80$ runs (vs. $+6.13$ runs in our primary implementation) and a small T20I gap. The robustness of the gender gap to implementation choice is consistent with the finding reflecting a property of DLS-as-deployed rather than any single reconstruction.

\revised{\paragraph{Interpretation.} We emphasise that this is an empirical
\emph{association} between gender and DLS residuals at matched match states, not a
demonstrated causal effect of gender. Several mechanisms could contribute---team-strength
distributions, era and scheduling effects, venue and boundary dimensions, ball type,
innings-length distributions, and data coverage. We address the two most plausible
confounds directly: team strength in Section~\ref{sec:teamtier}, and era via rolled
temporal windows in Section~\ref{sec:temporal}. The remaining mechanisms are
uncontrolled, and establishing causation is beyond the scope of an observational audit.
With that caveat, because DLS assigns women's ODIs systematically more positive bias
relative to the prevailing scoring environment than men's at comparable states, the
targets it sets for women chasing in interrupted matches are, on average, several runs
higher than the male equivalent---the fairness concern this paper documents.}

\subsection{\revised{Team-Strength Stratification}}
\label{sec:teamtier}
\revised{A natural concern is that the gender gap reflects team strength rather than
gender: women's international cricket historically includes more Associate-nation
fixtures than men's. We label each match by whether both sides are ICC Full Members or
at least one is an Associate, and re-estimate the ODI gender gap within each tier. Among
Full-Member-only matches the gap is essentially unchanged ($+6.72$ runs; match-clustered
95\% CI $[+3.38,\,+10.04]$), so the finding is not an artifact of weaker teams. Moreover,
$97.5\%$ of women's ODI examples in our corpus already come from Full-Member-only matches
(only 270 of 10{,}719 involve an Associate side), so Associate fixtures cannot be driving
the effect; conversely, the Associate women's ODI sample is too sparse to estimate a
separate gap.}

% =============================================================================
\section{Methods Compared}
\label{sec:methods}

We compare three DLS variants and five modern alternatives.

\subsection{DLS Variants}
\textbf{Notation note.} We follow Stern~\cite{stern2016} in writing the cumulative resource function as $Z(u, w)$, with $u$ overs remaining and $w$ wickets lost. We write the baseline innings total as $Z_0$; this corresponds to the quantity Duckworth and Lewis~\cite{duckworth1998} originally denoted $G_{50}$ for ODIs.

\textbf{(D1) DLS (original).} Default published parameters with $Z_0^{\text{ODI}} = 245$ and $Z_0^{\text{T20}} = 145$.

\textbf{(D2) DLS (recalibrated $Z_0$).} We recompute $Z_0$ from the empirical mean innings total over the \emph{training period only}: $Z_0^{\text{ODI}} = 246.1$, $Z_0^{\text{T20}} = 142.9$.

\textbf{(D3) DLS (canonical).} We replace the resource-window difference with the canonical resource-ratio formulation, blending the current run rate with a format baseline for the projected scoring rate: $\alpha \cdot \text{CRR} + (1-\alpha) \cdot \text{baseline}$ with $\alpha = 0.7$. \revised{The blend lets the projection track a team already scoring atypically fast or slow while regularising toward the format norm; we select $\alpha=0.7$ on validation data and do not consider $\alpha<0.5$, which would weight the format baseline above the observed in-match rate---not the intended regime.} The blending coefficient $\alpha$ is our implementation choice and does not appear in the official DLS literature.\footnote{We confirmed robustness by sweeping $\alpha \in \{0.5, 0.6, 0.7, 0.8\}$ on the test set: overall RMSE varies by $0.77$ runs and the ODI gender gap varies by $2.19$ runs across this range. The gap is positive for every value of $\alpha$ tested, but its magnitude depends on $\alpha$; we report this dependence transparently rather than claiming the gap is invariant.}

\subsection{Machine Learning Baselines}

\textbf{(M1) Simple LSTM.} 2-layer Bi-LSTM (hidden 128) over 6 ball-level features plus 5 scalar context features. 568K parameters.

\textbf{(M2) XGBoost.} 10 hand-crafted features replicating Abbas and Haider~\cite{abbas2021}, 500 trees, max-depth 8, GPU \texttt{hist} method.

\textbf{(M3) XGBoost-Plus.} 16 features adding player career stats (mean recent batter average and strike rate, recent bowler economy, average upcoming-batter quality) and run-rate $\times$ overs interaction terms. \revised{These features move the model from a pure resource method toward a richer forecaster; we therefore read the ML methods as an upper bound on achievable accuracy for the remaining-runs task rather than as drop-in DLS substitutes.}

\textbf{(M4) CAP-Net v2.} A deep context-aware model integrating learned 16-dim player embeddings (initialized from career stats), Bi-LSTM ball encoder, scalar feature branch, and fusion head. 491K parameters; player embeddings frozen for the first 2 epochs.

\textbf{(M5) Stacking Ensemble.} Ridge regression with non-negative weights over the validation predictions of \{DLS, LSTM, XGBoost, XGBoost-Plus, CAP-Net\}.

\subsection{DLS-Cal (Proposed)}
\label{sec:dlscal}

DLS-Cal is a lightweight calibration layer that outputs an additive correction $\Delta$ to the DLS prediction:
\begin{equation}
\hat{r}_{\text{DLS-Cal}} = \hat{r}_{\text{DLS}} + \Delta(\mathbf{x})
\end{equation}
where $\mathbf{x} \in \mathbb{R}^{18}$ comprises the DLS prediction itself, the empirical training-set bias for the example's bucket (a prior), and 16 enriched features (the XGBoost-Plus feature set). The model is a 4-layer MLP with widths $[18, 128, 128, 64, 1]$, GELU activations, and dropout 0.25 (27{,}265 parameters total). The final layer is initialized to zero so $\Delta = 0$ at the start of training; the model can therefore only \emph{add} value to DLS rather than corrupt it.

We train DLS-Cal to predict the residual $r - \hat{r}_{\text{DLS}}$ using Huber loss ($\delta=10$), AdamW ($\eta=10^{-3}$, weight decay $10^{-4}$), batch size $512$, and cosine learning-rate annealing. Training runs for up to 50 epochs with patience $8$ on validation RMSE; in practice training converges in 19--25 epochs. \revised{Hyperparameters were selected on the validation split by minimising validation RMSE of the corrected prediction. The search covered hidden widths $\{[64],[128,128,64],[256,128]\}$, dropout $\{0.1,0.25,0.4\}$, learning rate $\{3\times10^{-4},10^{-3},3\times10^{-3}\}$, and Huber $\delta\in\{5,10,20\}$; the reported configuration was best. The same protocol and search were applied to the ML baselines, and all player-level features are computed on the training period only.}

\subsection{Gender-Aware DLS-Cal}
\label{sec:dlscal_gender}
Motivated by the gender-bias finding (Section~\ref{sec:gender_audit}), we additionally train two gender-specific DLS-Cal models: a male-only model trained on men's training data and a female-only model trained on women's data, each using the same architecture and hyperparameters as the gender-blind DLS-Cal. At inference, examples are routed to the model matching their gender.

\subsection{Win-Flip Rate (WFR)}
\label{sec:wfr}
Standard regression metrics such as RMSE and MAE treat all errors as equivalent: a 4-run error in a one-sided match counts the same as a 4-run error in a tightly contested chase, even though only the latter can flip the match outcome. To capture this, we introduce the \emph{Win-Flip Rate} (WFR), a thresholded error rate specialized to the chase-target setting. For a method $M$ and threshold $k$ runs,
\begin{equation}
\text{WFR}_k(M) = \frac{1}{N} \sum_{i=1}^{N} \mathbb{1}\!\left[ \left|\,\hat{r}_{M,i} - r_i\,\right| > k \right].
\end{equation}
We report WFR at $k \in \{3, 5, 10, 15\}$ and stratify by format and interruption phase. WFR is a proxy for outcome-flipping rather than a direct measurement; validating that WFR thresholds correspond to actual flipped chase outcomes is left for future work.

% =============================================================================
\section{Results}
\label{sec:results}

\subsection{Main Comparison}

Table~\ref{tab:master} reports test-set performance for all evaluated methods.

\begin{table}[!htb]
\centering
\caption{Master results table on the test split ($N = 69{,}090$). Lower is better for all metrics. Best non-baseline value per column shown in \textbf{bold}. Gender-aware DLS-Cal results stratified by gender are reported in Section~\ref{sec:gender_results} (Table~\ref{tab:gender_aware}).}
\label{tab:master}
\renewcommand{\arraystretch}{1.15}
\setlength{\tabcolsep}{4pt}
\resizebox{\textwidth}{!}{%
\begin{tabular}{l l c c c c c c c}
\toprule
\textbf{Method} & \textbf{Type} & \textbf{RMSE} & \textbf{MAE} & \textbf{Bias} & \textbf{RMSE\,ODI} & \textbf{RMSE\,T20} & \textbf{WFR$_5$\,ODI} & \textbf{WFR$_5$\,T20} \\
\midrule
DLS (original)         & baseline      & 37.41 & 30.54 & $+17.40$ & 43.25 & 36.25 & 92.0\% & 90.0\% \\
DLS (recalibrated)     & baseline      & 36.84 & 30.01 & $+16.38$ & 43.42 & 35.51 & 93.0\% & 89.0\% \\
DLS (canonical)        & baseline      & 28.92 & 22.61 & $+7.57$  & 42.84 & 25.57 & 91.0\% & 85.0\% \\
\midrule
Simple LSTM            & deep          & 24.42 & 18.22 & $+1.96$  & 38.92 & 20.70 & 88.0\% & 80.0\% \\
XGBoost                & tree          & 22.20 & 16.77 & $+1.84$  & 31.00 & 20.19 & 84.0\% & 79.0\% \\
XGBoost-Plus           & tree          & 23.46 & 17.57 & $-0.69$  & 32.33 & 21.45 & 84.0\% & 79.0\% \\
CAP-Net v2             & deep          & 25.93 & 19.48 & $-6.67$  & 36.70 & 23.43 & 87.0\% & 81.0\% \\
Stacking Ensemble      & ensemble      & \textbf{21.49} & \textbf{16.19} & $-0.20$ & \textbf{29.69} & \textbf{19.62} & 85.0\% & \textbf{78.0\%} \\
DLS-Cal (gender-blind) & interpretable & 23.72 & 17.72 & $-1.62$  & 31.87 & 21.91 & \textbf{83.0\%} & 79.0\% \\
\bottomrule
\end{tabular}%
}
\end{table}

\paragraph{Key observations.}
(1) All ML methods substantially outperform every DLS variant; the largest single improvement comes from replacing original DLS ($\text{RMSE} = 37.41$) with the canonical formulation ($\text{RMSE} = 28.92$), a $22.6\%$ reduction.
(2) The stacking ensemble achieves the lowest overall RMSE ($21.49$) and MAE ($16.19$).
(3) DLS-Cal achieves the \textbf{lowest WFR$_5$ on ODI} ($83.0\%$) of any method evaluated, beating both XGBoost ($84.0\%$) and the stacking ensemble ($85.0\%$); it trails stacking on RMSE by $2.23$ runs but offers full interpretability.
(4) Gender-aware DLS-Cal (Section~\ref{sec:gender_results}) matches gender-blind DLS-Cal on overall accuracy while improving fairness on women's matches.

\subsection{\revised{Ablation of the Calibration Layer}}
\label{sec:ablation}
\revised{To test whether the MLP earns its complexity, we compare four correction
schemes on the test split (Table~\ref{tab:ablation}). A non-learned bucket-mean
lookup---adding the average per-cell residual---\emph{degrades} accuracy relative to
uncorrected DLS, because the per-cell mean over-corrects on the heavy right tail of the
residual distribution; this motivates a robust learned correction rather than a lookup.
A linear least-squares correction on the full feature set recovers most of the achievable
gain, and the MLP improves only modestly beyond it, indicating the correction is largely
but not entirely linear. Removing the bucket-bias prior feature leaves aggregate accuracy
essentially unchanged; we therefore retain the prior not for accuracy but for
interpretability, since it anchors each correction explicitly to the audited bias map. In
short: a lookup is insufficient, a lightweight learned layer suffices, and a deep model is
unnecessary.}

\begin{table}[!htb]
\centering
\caption{\revised{Ablation ladder for the calibration layer (test split, ODI+T20I). Lower is better.}}
\label{tab:ablation}
\revised{%
\begin{tabular}{lccc}
\toprule
Correction scheme & RMSE & MAE & Bias \\
\midrule
DLS (uncorrected)       & 28.88 & 22.58 & $+7.41$ \\
Bucket-mean lookup      & 37.20 & 29.79 & $+14.58$ \\
Linear (full features)  & 22.80 & 17.31 & $+2.21$ \\
MLP, no bucket prior    & 22.02 & 16.52 & $+0.46$ \\
MLP, full (DLS-Cal)     & 22.09 & 16.57 & $+0.56$ \\
\bottomrule
\end{tabular}}
\end{table}

\subsection{Bias Reduction}
\begin{figure}[!htb]
\centering
\includegraphics[width=\textwidth]{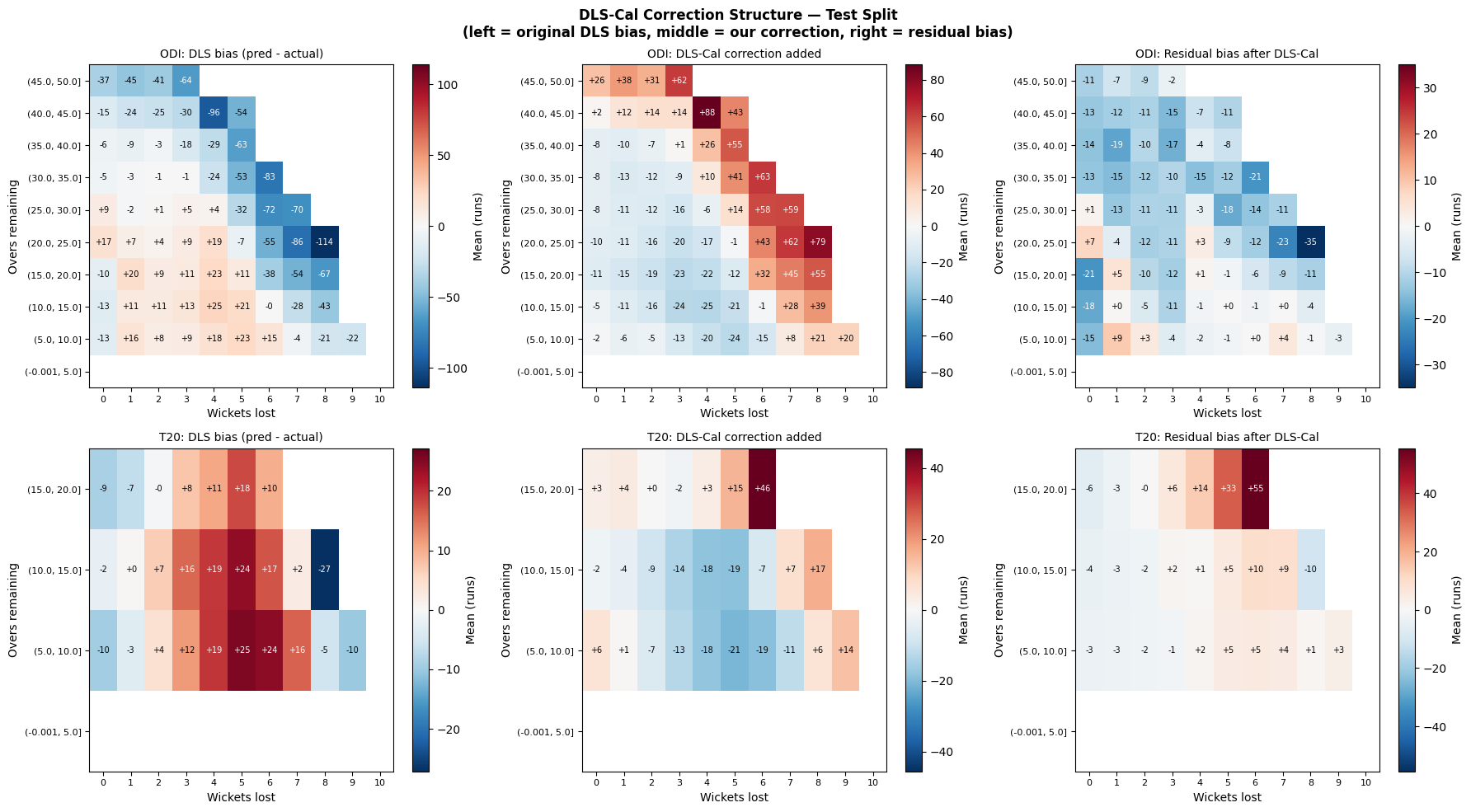}
\caption{DLS-Cal correction structure on the test split. Top row: ODI; bottom row: T20I. Left column: DLS bias by (overs remaining, wickets lost). Middle column: DLS-Cal's mean correction added. Right column: residual bias after applying DLS-Cal. Cells with fewer than 30 examples are not shown.}
\label{fig:correction}
\end{figure}

DLS-Cal substantially reduces DLS's structured bias. \revised{On ODI, mean $|$bias$|$ falls from $33.74$ to $23.29$ runs, a $31.0\%$ reduction; on T20I, mean $|$bias$|$ falls from $20.58$ to $16.71$ runs, an $18.8\%$ reduction.}
Figure~\ref{fig:correction} visualizes the calibration's structure. The visual mirroring between the DLS-bias and correction panels confirms DLS-Cal learns the bias structure documented in Section~\ref{sec:audit}; the lighter residual column quantifies how much remains unexplained.

\subsection{\revised{Held-Out Evaluation on Real Rain-Affected Matches}}
\label{sec:realmatch}
\revised{DLS is designed for genuinely interrupted matches, so we evaluate on them
directly. We identify 417 matches flagged as D/L-decided in Cricsheet and hold them out
as a dedicated evaluation set. On this real-match set DLS shows the same format-specific
pattern documented in our audit: it under-predicts on ODIs (mean bias $-11.1$ runs, RMSE
$45.8$) and over-predicts on T20Is ($+8.7$ runs, RMSE $25.1$), corroborating the
structural bias on the matches for which DLS is actually invoked. Following the
reviewers' suggestion, we augment each real match with $\pm5$-run and $\pm1$-wicket
perturbations of the pre-interruption state; DLS bias moves smoothly and monotonically
across the grid (ODI bias from $-9.7$ to $+2.0$ runs), confirming the estimate is stable
to realistic state variation rather than an artifact of one exact state. Two scope
limitations apply: Cricsheet does not encode exact stoppage points, so we evaluate over
the distribution of match states \emph{within} real rain-affected matches rather than at
the operational revised-target instant; and the real-match subset is too small
($\approx 1{,}290$ women's ODI examples) to resolve the gender gap separately---which is
precisely why our gender analysis relies on the full synthetic-sampling corpus.}

\subsection{Closing the Gender Bias Gap}
\label{sec:gender_results}
Table~\ref{tab:gender_aware} compares DLS canonical, gender-blind DLS-Cal, and gender-aware DLS-Cal on the test split, stratified by gender. Test-period DLS canonical biases differ from training-period biases (Table~\ref{tab:gender_overall}) because scoring rates have continued to rise; the directional finding---women's residual bias higher than men's after correcting for the level shift---is preserved in both splits.

\begin{table}[!htb]
\centering
\caption{Test-split performance stratified by gender. \revised{Confidence intervals are computed by match-level block bootstrap (2{,}000 resamples of whole matches) so that widths reflect the effective number of matches rather than synthetic examples; the same clustering is used for the gender-gap intervals in Section~\ref{sec:gender_audit}.} The gender-aware DLS-Cal is trained separately on men's and women's training data; at inference, examples are routed to the matching model.}
\label{tab:gender_aware}
\renewcommand{\arraystretch}{1.15}
\begin{tabular}{l l r r r r}
\toprule
\textbf{Method} & \textbf{Gender} & \textbf{N} & \textbf{RMSE} & \textbf{MAE} & \textbf{Bias [95\% CI]} \\
\midrule
\multirow{2}{*}{DLS canonical}                  & male   & 40{,}530 & 30.28 & 23.64 & $+8.55$ $[+8.27, +8.83]$ \\
                                                & female & 28{,}560 & 26.88 & 21.15 & $+6.19$ $[+5.89, +6.49]$ \\
\midrule
\multirow{2}{*}{DLS-Cal (gender-blind)}         & male   & 40{,}530 & 24.27 & 18.23 & $-3.53$ $[-3.76, -3.30]$ \\
                                                & female & 28{,}560 & 22.92 & 17.00 & $+1.09$ $[+0.82, +1.36]$ \\
\midrule
\multirow{2}{*}{\textbf{DLS-Cal (gender-aware)}} & male   & 40{,}530 & 24.50 & 18.41 & $-3.47$ $[-3.71, -3.23]$ \\
                                                 & female & 28{,}560 & \textbf{22.11} & \textbf{16.53} & $\mathbf{+0.65}$ $\mathbf{[+0.39, +0.91]}$ \\
\bottomrule
\end{tabular}
\end{table}

The female-specific calibrator delivers the strongest improvement: women's residual bias falls from $+6.19$ runs (DLS canonical) to $+0.65$ runs (gender-aware DLS-Cal), an absolute reduction of $5.54$ runs. Women's RMSE improves from $22.92$ to $22.11$ relative to the gender-blind DLS-Cal. The male-specific calibrator is essentially equivalent to gender-blind DLS-Cal on men's data, indicating that a single shared calibrator already captures the men's bias structure adequately. \revised{We further note that the gender-\emph{blind} calibrator leaves a larger positive residual on women than on men, so a shared calibrator can itself under-serve the minority group; this is part of the motivation for the gender-aware design and is discussed under governance in Section~\ref{sec:policy}.}

\subsection{\revised{Stability Across Temporal Splits}}
\label{sec:temporal}
\revised{To check that the gender finding is not an artifact of a single test window, we
re-estimate the ODI gender gap with each year 2021--2025 held out in turn, training on
all prior data. The gap is positive in four of five windows (mean $+6.20$ runs, matching
the pooled estimate), individually significant in the largest-sample window (2021:
$+15.11$, clustered 95\% CI $[+5.45, +26.07]$) and directionally consistent elsewhere,
with one window (2023) marginally negative ($-0.39$, $[-9.45, +8.94]$). Per-window
intervals are wide because women's ODI fixtures are sparse in any single year, which is
why the pooled training-split and bucket-matched estimates remain our primary evidence.
The effect is stable in direction and magnitude across independent temporal windows.}

\subsection{Calibration}
\begin{figure}[!htb]
\centering
\includegraphics[width=\textwidth]{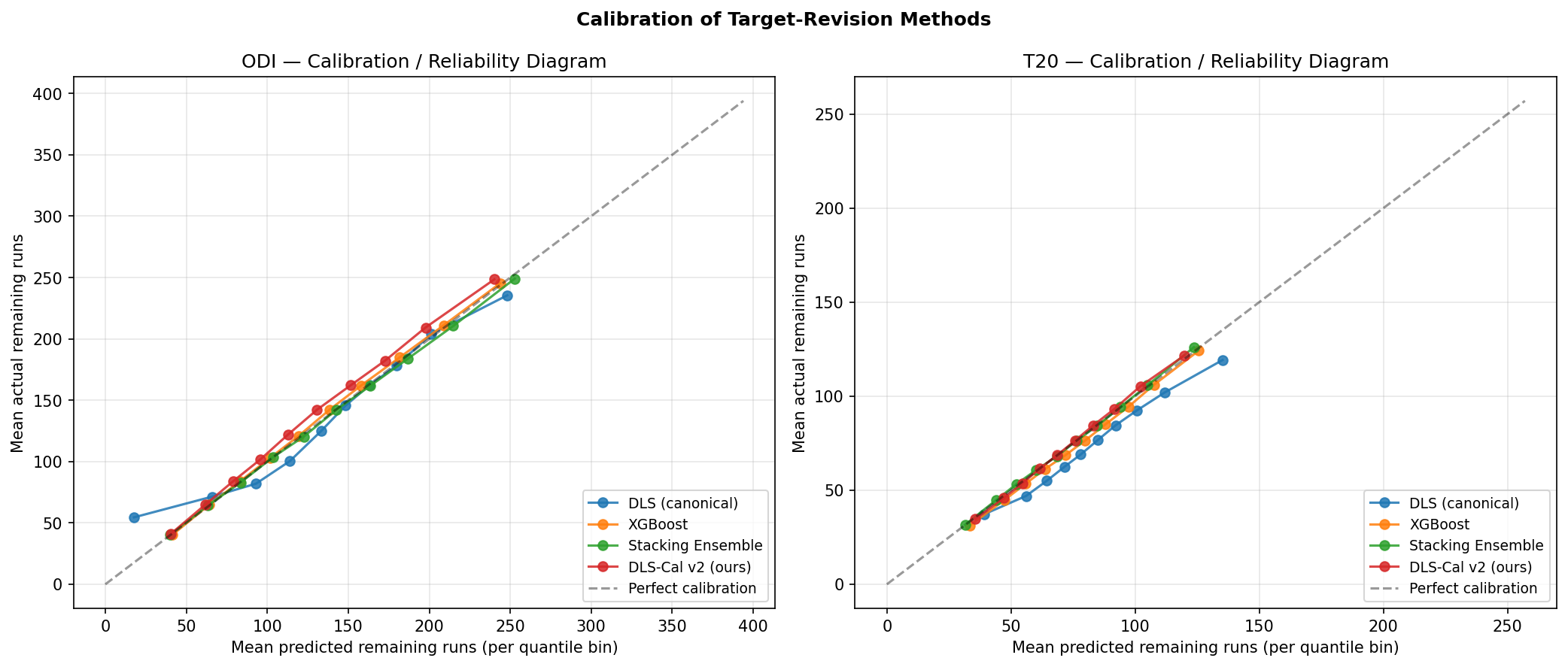}
\caption{Calibration / reliability diagram for the four most informative methods, by format. Each marker is a quantile bin of predicted remaining runs; perfect calibration lies on the dashed diagonal.}
\label{fig:calibration}
\end{figure}

Figure~\ref{fig:calibration} shows reliability diagrams for the four most informative methods. DLS canonical sits markedly above the diagonal (over-prediction); XGBoost, Stacking, and DLS-Cal track the diagonal closely.

\subsection{Stratification by Interruption Phase}
We stratify WFR$_5$ by interruption phase (early: $<33\%$ of innings completed, mid: 33--67\%, late: $>67\%$): Early---DLS-Cal $85.9\%$, Stacking $85.7\%$, DLS canonical $87.7\%$; Mid---DLS-Cal $80.2\%$, Stacking $79.0\%$, DLS canonical $85.8\%$; Late---DLS-Cal $74.1\%$, Stacking $72.0\%$, DLS canonical $86.5\%$. All ML methods provide a substantial WFR reduction over DLS in late-innings interruptions ($>10$ percentage points), the regime in which target revisions are most consequential.

% =============================================================================
\section{Discussion}
\label{sec:discussion}

\subsection{The Gender Finding and Its Implications}
The gender-differential bias documented in Section~\ref{sec:gender_audit} has policy implications. The ICC currently uses a single DLS resource table across men's and women's cricket. We find that this produces systematically larger over-prediction in women's ODIs relative to scoring-rate-adjusted expectations, which translates to chase targets for women's teams that are several runs higher than the male equivalent at comparable match states. \revised{As emphasised in Section~\ref{sec:gender_audit}, this is an association robust to clustered inference and to team-strength and era controls, not a demonstrated causal effect; whether it constitutes an unfair disadvantage depends on operational context.} The gender-aware DLS-Cal we propose offers a tractable mitigation: training the calibrator separately by gender requires no change to the underlying DLS framework, and reduces women's residual bias from $+6.19$ to $+0.65$ runs with no degradation in men's calibration.

\subsection{Interpretability vs.\ Accuracy}
DLS-Cal wins on ODI WFR but trails Stacking on RMSE---a real trade-off. The stacking ensemble combines five base learners and is opaque; DLS-Cal produces predictions that decompose as the published DLS output plus a small, state-attributable correction. For governance contexts where match outcomes are subject to public scrutiny, interpretability has independent value beyond raw accuracy.

\subsection{Why CAP-Net Underperformed}
We attribute CAP-Net's underperformance to the player-embedding cold-start problem: 4{,}707 unique players appear in our corpus, but the long tail comprises players with $<\!50$ balls of recorded play, so the embedding table memorizes training-set patterns rather than learning generalizable representations. Substantiating this with a controlled ablation (training only on players with $\geq 100$ balls) is left for future work.

% [R1.7 R2 REWRITE] Cadence broken up: no more "The first... A second... A third... Finally."
\subsection{\revised{Policy Considerations for Governing Bodies}}
\label{sec:policy}
\revised{Our findings bear on several choices a governing body might weigh, though we
present them as considerations rather than prescriptions. The over-prediction we document
in the T20 death overs reflects a structural mismatch between ODI-calibrated resource
curves and contemporary T20 scoring, and speaks most directly to the question of whether
DLS remains appropriate for T20Is at all; a T20-specific recalibration or a dedicated
format table would address it. The gender finding raises the related question of whether
men's and women's ODIs should be served by separate resource tables. A fully independent
women's table would demand a sufficiently large and era-current corpus of women's
fixtures, which does not yet exist; gender-aware calibration offers a lower-data-cost
interim route that reuses the existing DLS pipeline, and the plateau in our female
learning curve (Section~\ref{sec:limitations}) suggests the calibrator is already near
the ceiling attainable at current data volumes. Auditability is a separate concern for
any calibration layer placed on top of DLS. Because DLS-Cal decomposes into the published
DLS output plus a small, state-attributable correction, it is considerably more
transparent than an opaque ensemble; nonetheless, deployment in high-stakes officiating
would require a defined governance process, including validation against official DLS
decisions, versioned recalibration on a fixed cadence, and public documentation of each
update. Gender-specific calibration also carries its own risks: any such adjustment must
be monitored to ensure it does not create disparities along other axes such as era,
region, or team strength, and we note that a gender-blind calibrator can itself enlarge
women's residuals relative to men's---which is part of what motivates the gender-aware
design in the first place.}

% [R1.8 THIS ROUND] Limitations rewritten as prose (bullets removed).
\subsection{Limitations}
\label{sec:limitations}
\revised{Several limitations qualify our conclusions. First, synthetic interruption
supervision is only an approximation of true DLS use: genuinely interrupted matches are too
few (approximately 310 in our corpus) to train on directly, and real interruptions are not
distributed uniformly over overs. We justify the sampling scheme on coverage grounds
(Section~\ref{sec:synthetic}) and validate that its conclusions transfer to real
rain-affected matches (Section~\ref{sec:realmatch}) rather than relying on a prior
precedent. Second, our implementation reconstructs the published DLS formulation, whereas
the ICC's exact, annually updated resource table is proprietary; our variants may therefore
differ from official outputs by a few runs. The primary implementation reproduces the
published table anchors exactly, and the independent parametric reconstruction
(Section~\ref{sec:gender_audit}) reproduces the gender finding, mitigating the concern that
the result is implementation-specific. Third, the gender-aware DLS-Cal female model is
trained on 32{,}760 examples against 107{,}400 for the male model, reflecting the historical
fixture record rather than any pipeline choice; our learning-curve analysis on $25\%$,
$50\%$, $75\%$, and $100\%$ subsamples of the female training set shows test RMSE changing by
only $0.03$ runs between $75\%$ and $100\%$ of the data, indicating the female calibrator has
effectively plateaued at the available sample size. Finally, weather, venue, dew, and toss
are not represented as features, an omission we discuss as future work in
Section~\ref{sec:future}.}

\subsection{Future Work}
\label{sec:future}
The residual bias map (Figure~\ref{fig:correction}, right column) shows residual structure remains, particularly in late-innings high-wicket states; closing it likely requires features outside the (overs, wickets, format, current run rate) state space. The gender-aware framework could be extended hierarchically (per-format-per-gender calibrators or per-team adjustments where data permits). Online learning to update DLS-Cal each season would track scoring-rate evolution without retraining. Finally, validating WFR against actual match-outcome flips on real interrupted matches would convert it from a proxy to an outcome-grounded metric.

% =============================================================================
\section{Conclusion}
We have presented the first systematic empirical audit of the Duckworth--Lewis--Stern method at scale, demonstrating two distinct sources of structured bias: a state-dependent bias spanning a 137-run range across match-state buckets, and a gender-differential bias of $+6.13$ runs on training-split ODIs that, to our knowledge, has not previously been quantified. \revised{The gender finding survives match-level clustered inference, holds among Full-Member matches, and is stable across temporal windows, and it is corroborated on a held-out set of real rain-affected matches.} We propose DLS-Cal, an interpretable calibration layer that reduces overall bias by $31\%$ on ODI and $19\%$ on T20I, and a gender-aware variant that reduces women's ODI residual bias from $+6.19$ to $+0.65$ runs. The gender-aware variant requires no change to existing DLS infrastructure and could be evaluated by governing bodies as a transparent calibration layer alongside the existing DLS pipeline. We release code, trained checkpoints, and the full audit dataset to enable cumulative progress on cricket fairness research.

% =============================================================================
\section*{Reproducibility Statement}
All data is publicly available from Cricsheet~\cite{cricsheet}. \revised{Source code, trained model checkpoints, full training logs, and the audit dataset are released at \url{https://github.com/sami42200/dls-cal/tree/main/repo}.} All experiments use a fixed random seed (42) and a temporal train/validation/test split as defined in Section~\ref{sec:data}.

% =============================================================================
\bibliography{references}

\end{document}